\documentclass[final]{cvpr}

\usepackage{times}
\usepackage{epsfig}
\usepackage{graphicx}
\usepackage{amsmath}
\usepackage{amssymb}

\usepackage{bm}
\usepackage{enumitem}
\usepackage{multirow}
\usepackage{color}
\newcommand{\minisection}[1]{\vspace{0.04in} \noindent {\bf #1}\ \ }

\usepackage[pagebackref=true,breaklinks=true,colorlinks,bookmarks=false]{hyperref}

\def\cvprPaperID{49} % *** Enter the CVPR Paper ID here
\def\confYear{CVPR 2021}

\begin{document}

%%%%%%%%% TITLE
\title{Continual learning in cross-modal retrieval}

\author{Kai Wang$^{1}$, Luis Herranz$^{1}$, Joost van de Weijer$^1$\\

$^{1}$ Computer Vision Center, Universitat Autònoma de Barcelona, Barcelona, Spain\\
{\tt\small \{kwang,lherranz,joost\}@cvc.uab.es}
}

\maketitle

%%%%%%%%% ABSTRACT
\begin{abstract}
   Multimodal representations and continual learning are two areas closely related to human intelligence. The former considers the learning of shared representation spaces where information from different modalities can be compared and integrated (we focus on cross-modal retrieval between language and visual representations). The latter studies how to prevent forgetting a previously learned task when learning a new one. While humans excel in these two aspects, deep neural networks are still quite limited. In this paper, we propose a combination of both problems into a continual cross-modal retrieval setting, where we study how the catastrophic interference caused by new tasks impacts the embedding spaces and their cross-modal alignment required for effective retrieval. We propose a general framework that decouples the training, indexing and querying stages. We also identify and study different factors that may lead to forgetting, and propose tools to alleviate it. We found that the indexing stage pays an important role and that simply avoiding reindexing the database with updated embedding networks can lead to significant gains. We evaluated our methods in two image-text retrieval datasets, obtaining significant gains with respect to the fine tuning baseline.
\end{abstract}

%%%%%%%%% BODY TEXT

\section{Introduction}
\begin{figure}
	\includegraphics[width=\columnwidth]{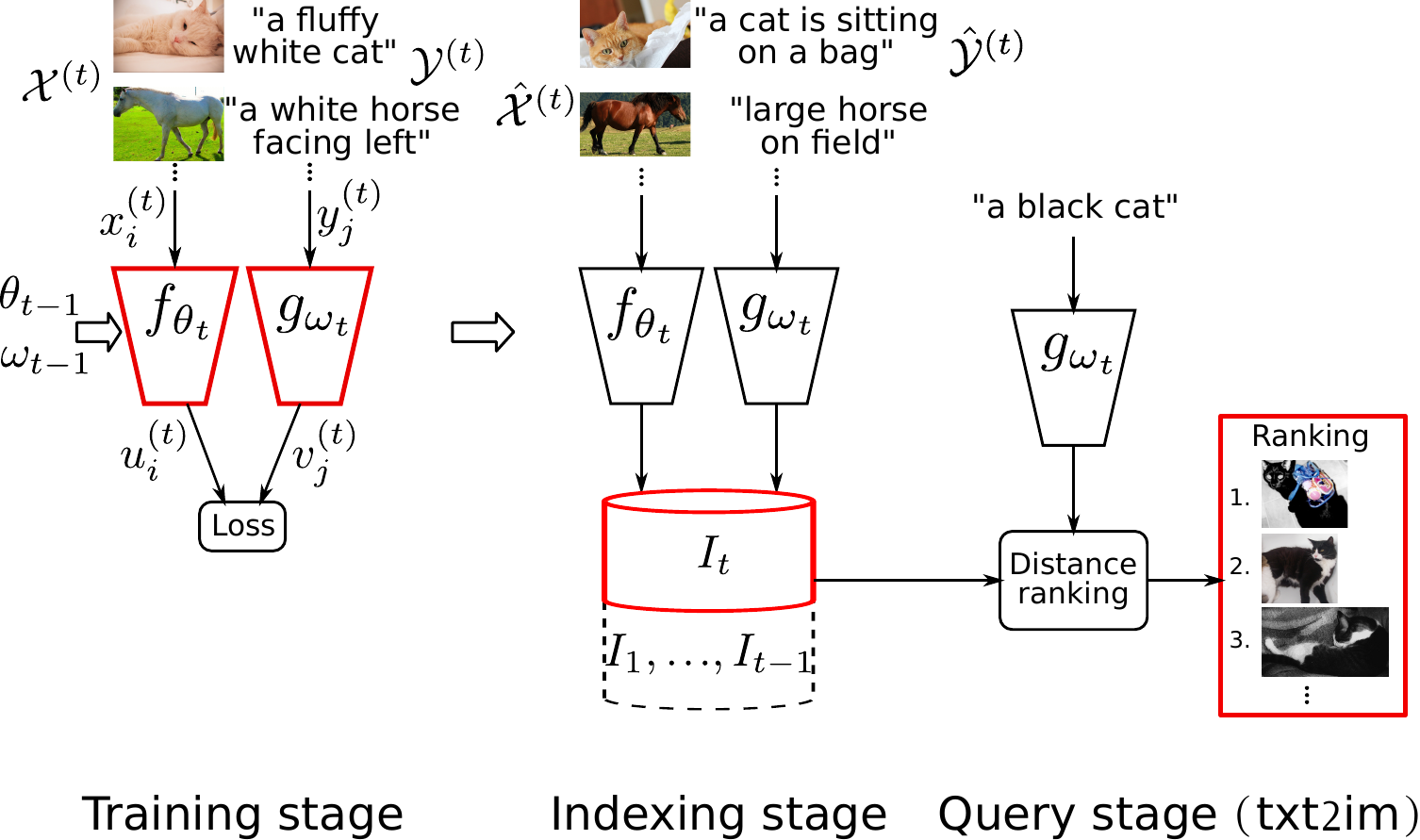}

	\caption{Stages in continual cross-modal retrieval (i.e. training feature extractors, indexing and query). The output of each stage is highlighted in red (i.e. feature extractors, index and ranking, respectively).\label{fig:retrieval_stages}}
\end{figure}
Human intelligence requires integrating, processing and comparing information from multiple modalities. Ideally, mental representations should lie in an abstract common space that is decoupled from the specific modality of the perceived information. Language and vision already interact in simple tasks such as object classification, where images are mapped to concepts in a closed vocabulary of categories. However, multimodal representations~\cite{baltruvsaitis2019multimodal} allow for richer interactions enabling cross-modal tasks such as cross-modal retrieval~\cite{chun2021probabilistic,wang2019learning,Chen_2020_CVPR,Wei_2020_CVPR,elhoseiny2017sherlock}, image captioning~\cite{Guo_2020_CVPR,Cornia_2020_CVPR,Pan_2020_CVPR}, visual question answering~\cite{niu2020counterfactual,Jiang_2020_CVPR,Chen_2020_CVPR_2,Wang_2020_CVPR}, and more recently text-to-image synthesis~\cite{li2019storygan,zhang2017stackgan}. Language models are also useful to extend visual classification beyond the limited categories seen during training by projecting to language spaces, also known as zero-shot recognition~\cite{frome2013devise,yang2020simple}.

Another characteristic of humans is their ability for continual learning, which allows us to perform well tasks learned long back in time. In contrast, neural networks suffer from catastrophic interference~\cite{mccloskey1989catastrophic,masana2020class}, which leads to almost complete forgetting of previous tasks when adapting to new ones, being a critical limitation to advance towards highly autonomous agents that can learn and adapt to changing environments. Continual learning (often referred to also as lifelong, sequential or incremental learning) in neural networks is an active research area, with recent methods addressing catastrophic forgetting with novel regularization~\cite{li2018learning,liu2018rotate,Yu_2020_CVPR}, architectural~\cite{masana2020ternary,berga2020disentanglement,serra2018overcoming} and (pseudo)-rehearsal~\cite{Liu_2020_CVPR_Workshops,shin2017continual,wu2018memory,wang2020bookworm} mechanisms. Most continual learning methods focus on classification tasks.

Motivated by these two challenges, here we study continual learning in multimodal embedding spaces applied to cross-modal retrieval, and the specific problems that arise in this scenario. In continual learning, training (of new tasks) can happen at different points in time. In a retrieval scenario, we must consider also the indexing operation, where an embedded representation is extracted from the input sample and stored in the database for future comparison. Since indexing in a continual setting could also happen at different points in time, we pay special attention to the role of this additional stage (see Fig.~\ref{fig:retrieval_stages}). An advantage of learning embedding networks instead of classification networks is that we operate in a single space shared by all tasks, so we can naturally retrieve data regardless whether we know or not the task related to that particular query sample (often referred to as task-aware and task-agnostic settings). Retrieval performance in continual learning is affected by how the embedded space may be distorted and cause representations to drift, as a result of the catastrophic interference. Additionally, these distortions and drifts may be unequal for each multimodality. Similarly, catastrophic forgetting affects differently to indexed data and query data.

In this paper we propose a continual cross-modal retrieval framework that can effectively perform retrieval in known and unknown domains. We identify and study the different factors that lead to forgetting in cross-modal embeddings and retrieval. Addressing those factors, we study modifications in the retrieval framework, network architecture and regularization that can help to alleviate them.

\section{Related Work}
\subsection{Deep metric learning}
%TODO. Siamese, Triplet, Negative mining

Deep metric learning learns both feature extraction and a distance metric in an end-to-end fashion. It maps images to an embedding space in which a simple distance metric such as the Euclidean distance can be applied. For training, it requires positive pairs (PP), which should be close in the embedding space, and negative pairs (NP), which are mapped at least a margin apart. Initial work was based on Siamese networks~\cite{bromley1994signature} which consist of two identical neural networks with shared weights, each taking one of the two inputs and map them to an embedding space. They are widely used in patch matching~\cite{simo2015discriminative}, face verification~\cite{schroff2015facenet}, image retrieval~\cite{gordo2016deep}, etc.  

Regarding the training loss, two of the most widely used are contrastive loss \cite{hadsell2006dimensionality} and triplet loss \cite{hoffer2015deep}. The former continually pushes similar instances closer, whereas negative pairs are only required to be at least a margin away. In the latter, similar samples are only required to be closer to each other than to any dissimilar ones. The training of Siamese and triplet networks is known to be difficult. Especially, since many of the negative pairs are already far apart in embedding space, they do not result in any training signal. Therefore, it was shown to be important to perform hard negative mining~\cite{simo2015discriminative}. Later works observed that it was computationally advantageous to first pass the images through a single network, and only form the pairs in the loss layer~\cite{oh2016deep,liu2017rankiqa}. Other losses include center loss~\cite{wen2016discriminative} and proxy-NCA~\cite{movshovitz2017no}.

\subsection{Cross-modal retrieval}
Cross-modal retrieval requires a coordinated representation~\cite{baltruvsaitis2019multimodal} that allows computing a similarity measure between the query representation and that of the retrieved data, even when they belong to different modalities and extracted with different feature extractors. There are two main aproaches to this problem: canonical correlation analysis (CCA)~\cite{hotelling1936cca} and metric learning~\cite{kulis2013metric}.

CCA~\cite{hotelling1936cca} learns linear projections to a space where the projections of two random vairables are maximally correlated, which makes it attractive to cross-modal retrieval. CCA has also been extended to deep networks~\cite{andrew2013deep,wang2015deep}, and in particular to cross-modal retrieval~\cite{feng2014cross,gong2014improving,klein2014fisher}. A limitation of CCA approaches is the expensive computation of the covariance matrix that requires having all data in memory.

Metric learning has also been applied sucessfully to cross-modal retrieval. Early examples of joint text-image embeddings are WSABIE~\cite{weston2011wsabie} and DeViSE~\cite{frome2013devise} which map image and text embeddings into a single space using ranking losses. Kiros \textit{et al.}~\cite{kiros2014unifying} applied a similar approach to sentences using an LSTM model. Socher \textit{et al.}~\cite{socher2014grounded} use an extended language model which includes dependency trees. Xu \textit{et al.}~\cite{xu2015jointly} propose a joint representation for video and sentences. Two-branch networks~\cite{wang2019learning} address image-text matching tasks with a bi-directional ranking loss. Multimodal representations have been also used for cross-modal retrieval of more structured visual-text documents, such as recipes~\cite{salvador2017learning,min2017being} and learning facts from images~\cite{elhoseiny2017sherlock}. 

Efficient retrieval from large databases is also a concern, so cross-modal hashing~\cite{bronstein2010data} learns compact representations in binary spaces where indexing and retrieval can be performed efficiently. Cross-modal hashing has also been extended to deep models~\cite{cao2016deep,jiang2017deep}.

\subsection{Continual learning}
A well known phenomenon in neural networks is catastrophic forgetting, where learning new tasks interferes with remembering previous ones~\cite{mccloskey1989catastrophic,masana2020class}. To enable networks to succeed in scenarios requiring continual learning, different techniques have been proposed.

A popular approach is to add regularization terms to the loss.  Weight regularization methods~\cite{kirkpatrick2017overcoming,zenke2017continual,aljundi2018memory,liu2018rotate} add quadratic terms to penalize large differences to the solution for previous tasks, weighted by some importance measure so differences in more important parameters are penalized more. Elastic weight consolidation (EWC)~\cite{kirkpatrick2017overcoming} uses the diagonal approximation of the Fisher information matrix to estimate the importance. Rotated EWC~\cite{liu2018rotate} proposes a reparametrization that makes EWC more effective. Synaptic intelligence (SI)~\cite{zenke2017continual} estimates the importance measure during training by accumulating gradients. Memory aware synapses (MAS)~\cite{aljundi2018memory} uses perturbation theory to estimate the importance in an unsupervised way. Forgetting can also be prevented by regularizing the activations, as in learning without forgetting (LwF)~\cite{li2018learning}, where a snapshot of the network right before starting to learn the new task  (and therefore not suffering interference from it) is used as a teacher and a distillation loss~\cite{hinton2015distilling} is used during the training of the new task. Encoder-based lifelong learning~\cite{rannen2017encoder} uses distillation in task-specific projections, estimated by autoencoders.

%- Rehearsal: Exemplars, Pseudo-rehearsal with pseudo-exemplars.
Another way of avoiding forgetting is rehearsal~\cite{robins1995catastrophic,rebuffi2017icarl}, where a fraction of data (i.e. exemplars) from previous tasks is kept and revisited during training, and pseudo-rehearsal~\cite{robins1995catastrophic,ans1997avoiding}, where pseudo-exemplars are sampled from an auxiliary model trained to model previous tasks.

Recent pseudo-rehearsal methods include deep generative models models~\cite{shin2017continual,wu2018memory}. Other approaches to continual learning include networks that expand their capacity to allocate new tasks~\cite{rusu2016progressive,yoon2017lifelong,schwarz2018progress} and task-attention mechanisms~\cite{serra2018overcoming}. 

While most works focus on classification, continual learning has also been studied in other settings such as image generation~\cite{wu2018memory,seff2017continual,nguyen2017variational}, word embeddings~\cite{kaji2017incremental,may2017streaming}, Atari games~\cite{kirkpatrick2017overcoming} and continual adaptation of agents~\cite{nagabandi2018deep}. MAS\cite{aljundi2018memory} is evaluated in facts learning that involves image and structured text. However, to our knowledge, there is not any work specifically studying the implications of continual learning in a retrieval setting, and catastrophic forgetting from the perspective of cross-modal embeddings.

\section{Continual cross-modal retrieval}
\subsection{Cross-modal deep metric learning}
Our framework is based on a two-branch network~\cite{wang2019learning}, with image-specific and text-specific embedding branches that project images and text into a common space. The image embedding operation is $u=f_{\bm{\theta}}\left(x\right)$, where $u\in\mathbb{R}^{E}$ is the image embedding of an input image $x$, extracted by the image embedding network $f_{\bm{\theta}}$, parametrized by $\bm{\theta}$. Similarly, the text embedding $v\in\mathbb{R}^{E}$ of an input text $y$ is obtained as $v=g_{\bm{\omega}}\left(y\right)$ by the text embedding network $g_{\bm{\omega}}$ parametrized by $\bm{\omega}$. Both $u$ and $v$ are normalized using $l_2$ norm. Images and text are compared in the embedding space using the Euclidean distance as $d\left(x,y\right)=\left\| u - v \right\|=\left\| f_{\bm{\theta}}\left(x\right) - g_{\bm{\omega}}\left(y\right) \right\|$.

The image set $\mathcal{X}=\left\lbrace x_i\right\rbrace_{i=1}^{N_\textnormal{I}} $  is aligned with a text set $\mathcal{Y}=\left\lbrace y_j\right\rbrace_{j=1}^{N_\textnormal{T}}$ via a pairwise similarity matrix $S$. This cross-modal pairwise similarity is indicated by a variable $s_{ij}\in S$ which takes value 1 when $x_i$ and $y_j$ are similar (i.e. \textit{positive pair}) and 0 otherwise (i.e. \textit{negative pair}). We want the distance between positive pairs to be significantly lower than the distance between negative pairs. In order to do that we use the bi-directional ranking loss of \cite{wang2019learning}, which selects triplets and imposes constraints
\begin{equation}
\begin{split}
	d\left(x_i,y_j\right)+m \le d\left(x_i,y_k\right)\\
	\text{s.t.} \ s_{ij}=1 \ \text{and}\ s_{ik}=0 
\end{split} \label{eq:triplet_anchor_image}
\end{equation}
and (in the other direction)
\begin{equation}
\begin{split}
	d\left(y_{i'},x_{j'}\right)+m \le d\left(y_{i'},x_{k'}\right)\\
\text{s.t.} \ s_{i'j'}=1 \ \text{and}\ s_{i'k'}=0 
\end{split}\label{eq:triplet_anchor_text}
\end{equation}
where $m$ is the predefined margin. The triplets are constructed based on a positive pair, and a negative pair creating by replacing either the image or the text by a dissimilar one. These triplet constraints are included using a margin-based loss function (where $\left[z\right]_+=\text{max}\left(0,z\right)$):
\begin{equation}
\begin{split}
	L_{\textnormal{T}}\left(\mathcal{X},\mathcal{Y}\right)=\lambda_1\sum_{i,j,k}\left[d\left(x_i,y_j\right)+m - d\left(x_i,y_k\right)\right]_+ \\
	+\lambda_2\sum_{i',j',k'}\left[d\left(y_{i'},x_{j'}\right)+m - d\left(y_{i'},x_{k'}\right)\right]_+
\end{split} \label{eq:ranking_loss}
\end{equation}

\subsection{Training, indexing and query stages}
In general, machine learning assumes two different stages, namely \textit{training} and \textit{evaluation} (or \textit{test}), which take place in that exact order (although in continual learning it is not the case). We focus on retrieval with a learned feature extractor (i.e. embedding networks in our case). In this scenario we identify three stages (see Fig.~\ref{fig:retrieval_stages}): 

% \begin{itemize}
%     \item 
    \textbf{Training (feature extractors(s))}. Described in the previous section, the training stage learns the embedding networks from the image and text datasets $\mathcal{X}$ and $\mathcal{Y}$, and its result is the parameters $\bm{\theta}$ and $\bm{\omega}$.
    
    % \item 
    \textbf{Indexing (database data)}. The database datasets $\hat{\mathcal{X}}$ and $\hat{\mathcal{Y}}$ to be indexed are processed using the embedding networks to obtain the text and image embeddings, which are subsequentially indexed in the database. Note that training data and database data are not required to be the same.
    
    % \item 
    \textbf{Querying (query data)}. This stage computes the similarity between a query sample and the indexed data. The result is a ranking with the most similar sample on top. In our cross-modal case there are two directions: querying with images, retrieving from indexed texts (\textit{im2txt}) and querying with text, retrieving from indexed images (\textit{txt2im}). 
    
% \end{itemize}

Note that these three stages are assumed to take place in that particular order, and a deployed system only performs the querying stage. For simplicity we consider that the database data is also used as training data, i.e.  $\hat{\mathcal{X}}=\mathcal{X}$ and $\hat{\mathcal{Y}}=\mathcal{Y}$.

\subsection{A framework for continual retrieval}
Now we consider a continual learning setting, in which data is presented as a sequence of tasks $\left\{\mathcal{T}^{(t)}\right\}^T_{t=1}$. Each task $\mathcal{T}^{(t)}=\left(\mathcal{X}^{(t)},\mathcal{Y}^{(t)},\mathcal{S}^{(t)}\right)$ involves data from a different domain (e.g. animals, vehicles). We assume that the embedding networks are updated (i.e. fine tuned) with data of a particular task (i.e. training stage) before indexing data of that task. The resulting parameters after training task $t$ are $\bm{\theta}_t$ and $\bm{\omega}_t$. 

\begin{figure}
	\includegraphics[width=\columnwidth]{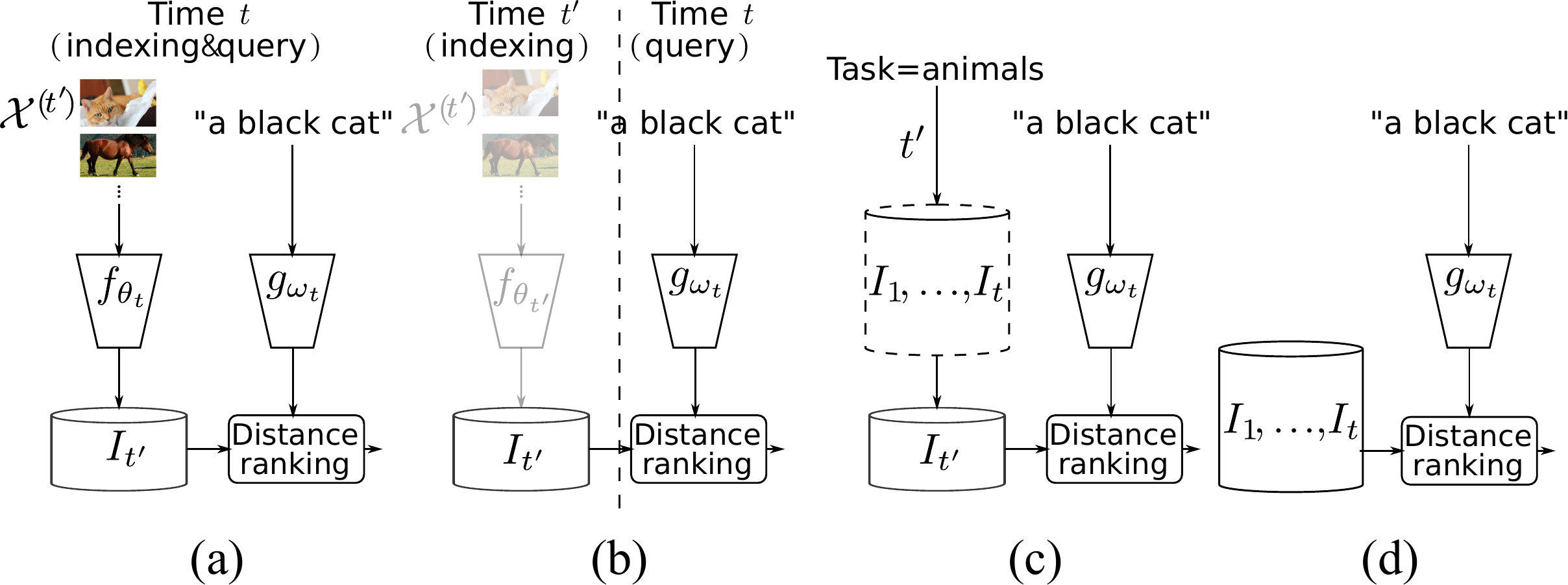}

	\caption{Variants of indexing data from a previous task $t'$ when queried at time $t>t'$ (a-b) and retrieval (c-d): (a) reindexing, (b) not reindexing, (c) task known, (d) task unknown.}\label{fig:retrieval_variants}
\end{figure}

The retrieval system is evaluated in the querying stage with separate data from every task. We consider two settings for evaluation: \textit{known task} and \textit{unknown task}, depending on whether that information is available at query time (see Fig.~\ref{fig:retrieval_variants}a-b). 
% The former setting is easier, since the knowledge of which task is the relevant one dramatically reduces the search space.

\begin{figure}
	\includegraphics[width=\columnwidth]{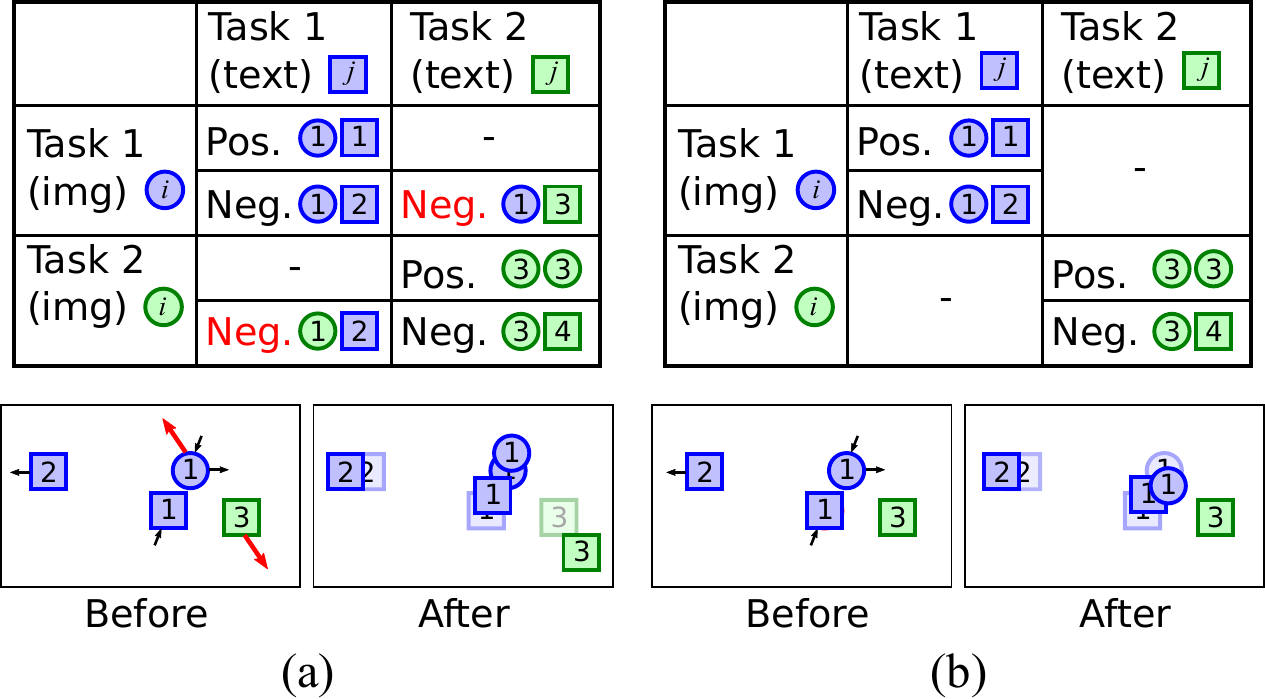}

	\caption{Types of pairs in continual cross-modal retrieval: (a) available in joint training, and (b) available in continual learning, i.e. without cross-task negative pairs (CTNP). CTNPs are crucial to avoid overlap between samples of different tasks (bottom). Best viewed in color.\label{fig:types_pairs}}
\end{figure}

As described previously, the network is trained using cross-modal positive and negative pairs. When all data is presented jointly, all negative pairs are available for sampling. However, in the continual setting, pairs are formed within the same task, i.e. combining samples from $\mathcal{X}^{(t)}$ and $\mathcal{Y}^{(t)}$. Thus, we further classify a negative pair $\left(x_i,y_j\right)$ as intra-task negative pair (ITNP), when $x_i\in\mathcal{X}^{(t)}$ and $y_j\in\mathcal{Y}^{(t)}$ ($s_{ij}=0$, $s_{ij}\in \mathcal{S}^{(t)}$), or as cross-task negative pair (CTNP), when $x_i\in\mathcal{X}^{(t')}$ and $y_j\in\mathcal{Y}^{(t)}$, $t'\neq t$. Note that, for simplicity, we assume that all positive pairs are intra-task. In continual retrieval, CNTPs are not available during training (see Fig.~\ref{fig:types_pairs}).

\subsection{Do or do not reindex?}
The conventional retrieval scenario assumes that training and indexing are performed once. In this case, there is only a static set of embeddings, extracted with the same network at the same time. The same network is used to extract embeddings from queries. In continual retrieval this may not be the case, since training and indexing are performed every time a new task is presented.

\textbf{Reindexing.} We first consider the straightforward extension of cross-modal retrieval that assumes that current and previous tasks all are reindexed with the version of the embedding networks with updated parameters $f_{\bm{\theta}_t}$ and $g_{\bm{\omega}_t}$ after a new task $t$ is learned (see Fig.~\ref{fig:retrieval_variants}a). We refer to this case as \textit{reindexing}. However, it has the drawbacks of being time and resource consuming, since it requires indexing the same data multiple times, and always requiring access to the image and text samples of previous tasks. It has the advantage that database and query samples are processed with the same networks.  

\textbf{No reindexing.} We also propose the variant \textit{no reindexing} that only indexes the data of current task $t$ after training task $t$ (see Fig.~\ref{fig:retrieval_variants}). This variant is more efficient, since database samples are processed only once, and flexible since it does not require access to previous images and text (only to their indexed embeddings for retrieval). On the other hand, no reindexing introduces asymmetry, since query embeddings are extracted with $f_{\bm{\theta}_t}$ (or $g_{\bm{\omega}_t}$), while database embeddings with $g_{\bm{\omega}_{t'}}$ (or $f_{\bm{\theta}_{t'}}$, with $t'\leq t$).

\section{Catastrophic forgetting in cross-modal embeddings}

\begin{figure}
	\includegraphics[width=\columnwidth]{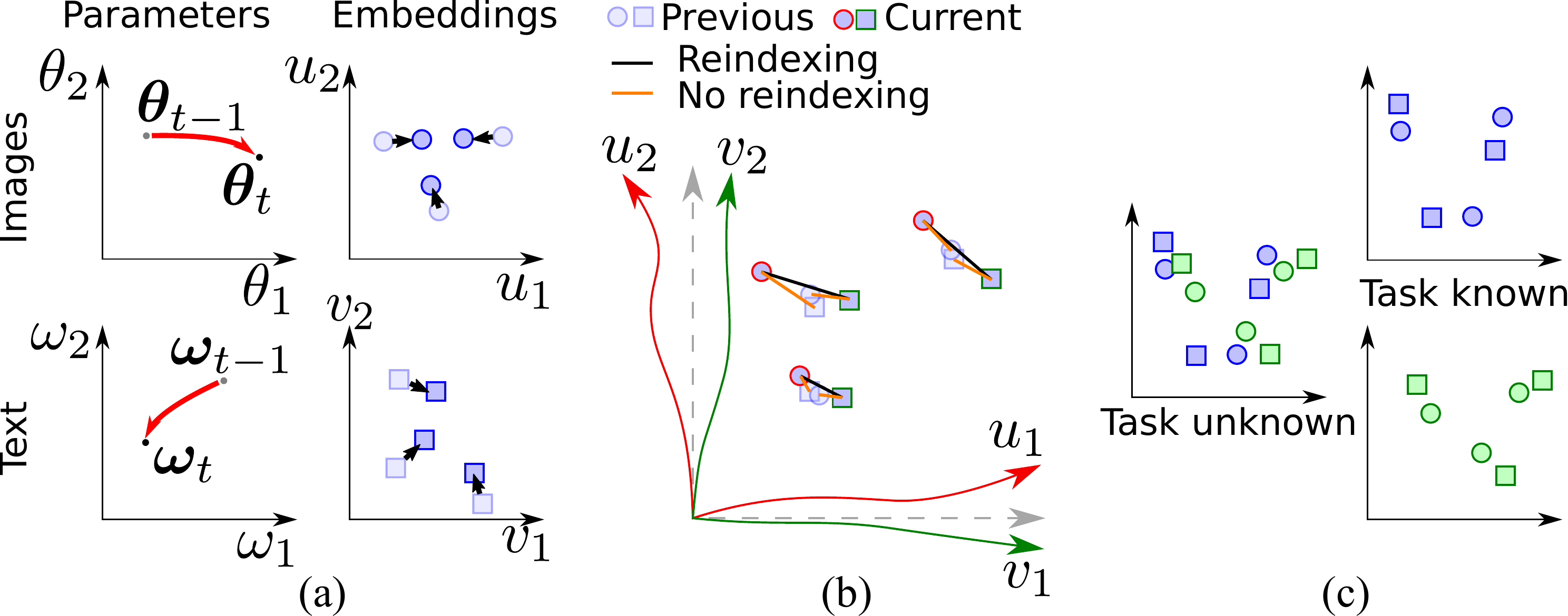}

	\caption{Causes of forgetting in cross-modal embeddings: (a) embedding networks become less discriminative due to drift in parameter space, and (b) unequal drift increases cross-modal misalignment, and (c) task overlap in embedded space (when task is unknown). Best viewed in color.\label{fig:causes_forgetting}}
\end{figure}

Learning a new task implies that the values of the network parameters will shift away from the previous ones. This is particularly important when the new task is very different from previous, causing interference between new and previous tasks that leads to lower performance in the latter. For simplicity, we will refer to this drop in performance as \textit{catastrophic forgetting}. In the following, we identify several phenomena that may lead to forgetting in  continual cross-modal retrieval.

\minisection{Embedding networks.}
We first consider forgetting in each embedding network separately, without considering pairwise interactions. As their parameters move away from the optimal values for $t-1$ (see Fig.~\ref{fig:causes_forgetting}a), the embeddings $u$ and $v$ will also drift from their previous values. In general, the new values $f_{\bm{\theta}_t}$ and $g_{\bm{\omega}_t}$ are less discriminative than previous previous $f_{\bm{\theta}_{t-1}}$ and $g_{\bm{\omega}_{t-1}}$, causing lower performance, because the embedding spaces of $u^{(t)}$ and $v^{(t)}$ are also less discriminative.

\minisection{Embedding misalignment.}
In the particular case of cross-modal networks, embeddings of different modalities may drift differently (see Fig.~\ref{fig:causes_forgetting}a). This unequal drift in $u$ and $v$ spaces causes additional misalignment that also leads to higher distances than in the optimal case. Note that unimodal retrieval with Siamese networks or Triplet networks does not suffer from this problem, since parameters are shared across de various branches.

\minisection{Task overlap.}
Negative pairs pull dissimilar samples away in the embedded space. However, in continual retrieval CTNPs cannot be sampled (unless we include some samples from previous tasks). CNTPs are the only repulsive force between samples of different tasks. Without them, it is likely that samples from different tasks will overlap in the embedded space (see Fig.~\ref{fig:causes_forgetting}a). Knowing the task at query time makes this problem less important, since data from other tasks are not considered at query time.

\section{Preventing forgetting}
In the following we propose several tools to alleviate forgetting by addressing the previous causes.

\subsection{Preventing embedding drift}

A common approach to prevent forgetting is regularizing the weights with a quadratic term in the loss that penalizes the weighted Euclidean distance (in the parameter spaces) to the solution for previous tasks~\cite{kirkpatrick2017overcoming,zenke2017continual,aljundi2018memory,liu2018rotate}. This can help to avoid significant drift in the embeddings and to keep them discriminative for previous tasks. We can write the particular regularization term for our case as
\begin{equation}
\begin{split}
    L_{\textnormal{R}}=\sum_k \Theta^{(t-1)}_k\left(\theta^{(t-1)}_k\!-\!\theta_k \right )^2 + \\ +\sum_{k'} \Omega^{(t-1)}_{k'}\left(\omega^{(t-1)}_{k'}\!-\!\omega_{k'} \right )^2 \label{eq:reg_EMB}
    \end{split}
\end{equation}
where $\Theta_k$ and $\Omega_{k'}$ control the regularization strength depending on the importance of $\bm{\theta_k}$ and $\bm{\omega_k}$, respectively, for previous tasks. During the training of the first task there is no regularization, i.e. $\Theta^{(0)}_k=0$ and $\Omega^{(0)}_{k'}=0$. The way to compute the importance differs in different methods. We consider two variants:

% \begin{itemize}
%     \item 
    \textbf{Global}. Here we estimate the importance with respect to the loss, adapting elastic weight consolidation (EWC) to our particular triplet loss as ($L_{\textnormal{TR}}$ represents the triplet loss): 
    
    \begin{equation}
        \Theta^{(t)}_k=\mathbb{E}_{x,y}\left [ \left ( \frac{\partial}{\partial \theta_k } L_{\textnormal{TR}}\left(\mathcal{X}^{(t)},\mathcal{Y}^{(t)} \vert \bm{\theta}_{t},\bm{\omega}_{t}\right) \right )^2 \right ]
    \end{equation} 

    which is computed by sampling triplets as in \ref{eq:triplet_anchor_image} and \ref{eq:triplet_anchor_text}, and analogously for $\Omega_{k'}$. This loss already takes into account triplets and their interactions.\\
    
    % \item 
    \textbf{Branch}. Instead of estimating importance values that depend on a joint loss, we consider regularizing each branch independently. In this case we estimate the importance using the approach memory aware synapses (MAS), which can be computed unsupervisedly for each branch with images or text. The importance for the image branch is estimated as:

    \begin{equation}
        \Theta^{(t)}_k=\Theta^{(t-1)}_k+\mathbb{E}_{x_i \sim \mathcal{X}^{(t)}} \left [ \frac{\partial } {\partial \bm{\theta}_k} l^2_2 (f_{{\theta}_t} (x_i)) \right] 
    \end{equation} 

which is accumulated over previously computed one. For the text branch the estimation of $\Omega_{k'}$ is analogous. In this equation, $l^2_2$ is the squared $l^2$ norm of the function outputs, which is used to estimate the importance of parameters in MAS method.

The final loss combines (\ref{eq:ranking_loss}) and (\ref{eq:reg_EMB}) as $L=L_{\textnormal{T}}+\lambda_3 L_{\textnormal{R}}$.

\subsection{Preventing unequal drift}

In order to prevent unequal drift we propose tying the networks by sharing layers at the top (bottom layers must remain modality-specific). In this way, the unequal drift can be alleviated since the gradients are tied and only differ in the lower layers.  

In some cases when the drifts in text and image embedding are in opposite directions, refraining from reindexing the database can be an effective tool to alleviate drift, since only one of the embeddings is affected while the other remains fixed. Fig.~\ref{fig:causes_forgetting}b illustrates how in that case no reindexing keeps matching pairs at lower distances.

\subsection{Decoupling retrieval directions}

So far we assumed only a single model is trained to perform both text to image and image to text retrieval. This is reasonable when embeddings are reindexed since the architecture and the loss are symmetric. However, when database data is not reindexed and query is, the forgetting is asymmetric. In that case we can decouple both directions and train one model for each direction, only regularizing the weights in the query branch. This can also be beneficial in some cases when the image and text embeddings drift in different directions, keeping one fixed in the previous position can keep the distance lower (see example in Fig.~\ref{fig:causes_forgetting}b).

\subsection{Preventing cross-task overlap}
The lack of CTNPs can lead to cross-task overlap, since there is no force separating them. However, reducing the drift, keeping the embeddings discriminative via weight regularization and sharing layers may indirectly help to keep tasks separated (we observed that in our experiments). 

Nevertheless, we made some preliminary experiments creating pseudo-CTNPs $\left(u_i,x_j\right)$ in models with decoupled retrieval directions using the already indexed embeddings (analogously for text to image retrieval for the other direction), but we found they did not help in our experiments, probably because the asymmetric force that only pushes the embeddings of one branch. In this case the gradients are only backpropagated through one branch. We leave their study more in depth for future work.

\section{Experiments}
\minisection{Baselines and variants}
We evaluate the different variants of our continual cross-modal framework in two tasks involving images and text, one focusing on regions and the other on scenes. We follow the implementation of the two-branch networks in~\cite{wang2019learning} where 4096-dim image features are extracted from a VGG-19 model trained on ImageNet, and text features are 6000-dim from HGLMM features (reduced with PCA from initial 18000-dim)~\cite{klein2014fisher}. The image branch includes two additional fully connected layers with sizes 2048 and 64 (for SeViGe, 2048 and 512 for SeCOCO) on top and $l_2$ normalization, and the same for the text branch. We focus our study on the two fully connected layers on top, while the initial feature extractors remain fixed. As in \cite{wang2019learning} we set $\lambda_1=1.0$, $\lambda_2=1.5$ and the margin $m=0.05$. The resulting model is trained with Adam\cite{kingma2014adam} and a learning rate of 0.0001, and using dropout after ReLu with probability 0.5. We evaluate different variations of this architecture:
{}
\begin{itemize}[noitemsep]
    \item \textbf{Joint vs continual.} We compare the variants of the proposed framework (\textit{continual}) with two baselines that learn all tasks jointly (\textit{joint}), differing on whether CTNPs are sampled or not during training.
    \item \textbf{Retrieval direction.} We evaluate both text to image retrieval (\textit{txt2im}) and image to text retrieval (\textit{im2txt}).
    \item \textbf{Task knowledge.} We evaluate both the cases where the task is \textit{known} and \textit{unknown}.
    \item \textbf{Reindexing.} We consider the embeddings for database samples are extracted when the corresponding task was learned (\textit{no reindex}), or are at the same time as the query embeddings (\textit{reindex}).
    \item \textbf{Weight regularization.} We consider fine tuning with no regularization (\textit{ft}), with joint regularization on the loss (\textit{EWC}) and with regularization on each $u$ and $v$ embedding independently (\textit{MAS}). We set $\lambda_3=10^6$.
    \item \textbf{Decoupled directions.} For \textit{no reindex} we also consider variants where EWC or MAS are only computed in the branch extracting query embeddings (e.g. \textit{MAS-txt} when MAS is computed only on the text branch). In this case we run two different experiments, each specialized for one particular retrieval direction.
    \item \textbf{Layer sharing.} We consider keeping both embedding networks independent (\textit{no sharing}) or sharing the top fully connected layer (\textit{sharing}).
    
\end{itemize}

We consider experiments where each task consist in updating the embedding networks by learning a new domain. After training the model, the same training data is indexed (i.e. we extract image and text embeddings) and then the retrieval performance can be evaluated. We report the final results after all tasks are learned. We use Recall@K as evaluation metric (with $K=10$, results for other $K$ in the supplementary material), with respect to the indexed data of the same domain (\textit{known}) or to the whole indexed data with all domains (\textit{unknown}). We repeat each experiment five times and report the average.

\subsection{Sequential Visual Genome}
\minisection{Sequential Visual Genome (SeViGe) dataset.} We created a dataset based on the regions with object-description pairs in the Visual Genome dataset~\cite{krishna2017visual}. Based on the object categories of those regions, we selected pairs related with the domains \textit{animals} (9 categories), \textit{vehicles} (6 categories) and \textit{clothes} (6 categories), which are learned in sequence as tasks in our experiments. Each task has a total of 10481, 7531 and 10200 training images, respectively, and additional 900/900, 600/600 and 600/600 for validation/test, respectively (100/100 per category).

\minisection{Cross-modal retrieval.}
We evaluate the different methods in cross-modal region image-text retrieval. The results for both directions are shown in Table~\ref{tab:seqvisualgenome}. We focus on \textit{average} for evaluation when the task is known, and \textit{A+V+C} to evaluate when the task is unknown (i.e. the aggregate of all domains). We first observe that training all tasks jointly delivers higher performance than the continual setting, as expected. Significant part of that superiority is due to CTNP, since the performance of joint training drops significantly when not sampled. This provides a more realistic and tighter upper bound, since in the continual scenario CTNPs are not available. This drop is more moderate when the task is known ($\sim$1-2\% known, $\sim$3-4\% unknown), since task overlapping is not a problem. This drop still suggests that CTNPs still contribute to shape the embedding spaces to be discriminative beyond simply avoiding task overlap. When the top layer is shared, the drop is also smaller, although the overall performance is also lower than not sharing the layer.

Focusing on continual learning we observe that the single modification that most reduces forgetting is not reindexing the database, which provides 3\% and 5.1\% boosts in \textit{im2txt} and \textit{txt2im} directions, respectively (1.6\% and 3.4\% if task is unknown). This surprising result, showing that reindexing can be harmful, suggests that the misalignment caused by the unequal drift of image and text embeddings is more critical than the misalignment caused by not extracting embeddings at the same time, and that keeping good and discriminative representations in the database is also important (recall that in \textit{no reindexing} only the query embedding has endured catastrophic interference). Note that this is specific to cross-modal retrieval because branches do not share parameters. This may not be the case in image-to-image or text-to-text retrieval with Siamese or Triplet networks because the embeddings of the two branches drift equally.

Sharing layers by itself gives an improvement of 1.4\% in \textit{im2txt}, while not having impact in \textit{txt2im}. Not reindexing also gives a similar boost as in the previous case.  Interestingly, sharing layers harms performance in joint training, while for the continual setting it improves the performance, probably because it reduces the unequal drift by tying the drift of both modalities at least in the shared layers.

Weight regularization has moderate impact and could harm the performance sometimes. Decoupling both modalities and applying regularization only in the query network extractor seems to help in some cases (e.g. +1.3\%/+1.6\% gain with \textit{MAS-txt} vs \textit{MAS} in \textit{txt2im}, and +0.7\%/+0.9\% in \textit{im2txt} in the \textit{sharing} architecture). In this dataset regularizing embeddings independently with MAS instead of the whole network with EWC seems to work better, although the differences are very marginal. Here we can see that the forgetting in embedding network is not a significant problem in cross-modal retrieval setting.

Overall, the best combination provides improvements of 6\%/6.3\% in known/unknown \textit{txt2im} retrieval, and more moderate improvements of 2.9\%/2\% in \textit{im2txt} retrieval.

\begin{figure}
\centering
	\includegraphics[width=\columnwidth]{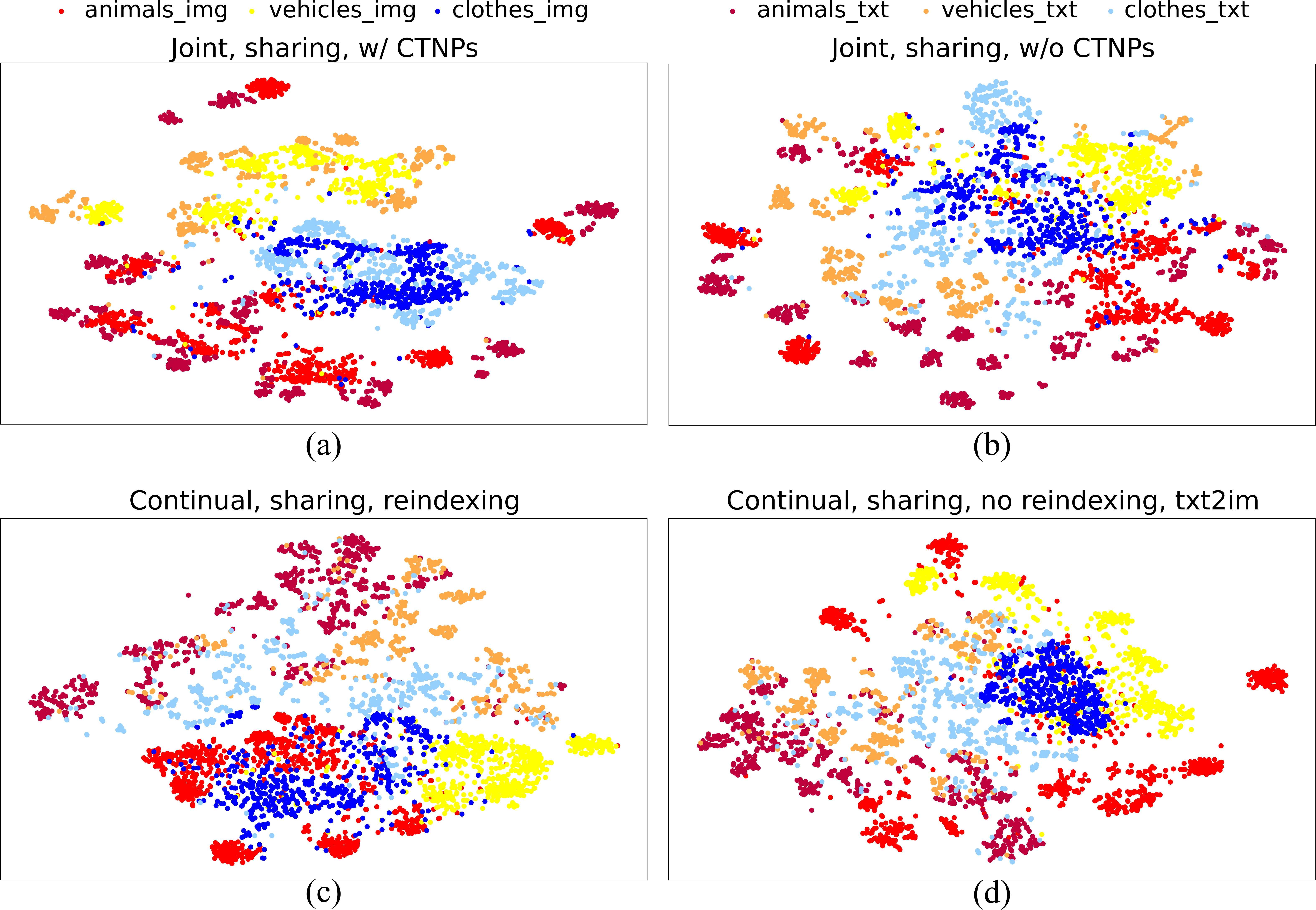}
	\caption{t-SNE visualization of the cross-modal embedding space of SeViGe, with the \textit{sharing} architecture: (a) joint training (with CTNPs), (b) joint training (without CTNPs), (c) continual (reindexing), and (d) continual (no reindexing). Best viewed in color.\label{fig:tsne_sevige}}
\end{figure}

\minisection{Insights about the embedding space.} 
We use t-SNE~\cite{maaten2008visualizing} to visualize the embedding space of variants with shared layers. Although distances in t-SNE do not reflect real distances, it is useful to identify structure. We combine text and image embeddings and runt-SNE, color coding data with modality and task labels. Joint training (see Fig.~\ref{fig:tsne_sevige}a) generates embeddings where data is structured clearly in separated tasks, and within tasks, in separated clusters (probably the categories within each task in SeViGe). This happens in both modalities, which also overlap, aligned according to the related clusters. Not sampling CTNPs still results in intra-task structure (see Fig.~\ref{fig:tsne_sevige}b, e.g. category clusters are clear), but the modalities are significantly more misaligned and with larger overlap, showing the important role of CTNPs in aligning modalities and separating tasks.  When learned in a continual fashion (see Fig.~\ref{fig:tsne_sevige}c), the misalignment is more extreme, even resulting in text and image samples distributed in different halves of the space. No reindexing (see Fig.~\ref{fig:tsne_sevige}d for \textit{txt2im} direction) seems to keep the image embeddings (database) more discriminative, which may explain the improved results compared to reindexing. For example, the image embeddings of animals and vehicles seem much better separated in Fig.~\ref{fig:tsne_sevige}d than in Fig.~\ref{fig:tsne_sevige}c.

\begin{table*}[t]
\centering
\def\arraystretch{1.0}
\setlength\tabcolsep{2pt}
\resizebox{\textwidth}{!}{%
\begin{tabular}{c||cc|ccc|ccccc||cc|ccc|ccccc}
\hline
\multirow{4}{*}{Domain} & \multicolumn{10}{c||}{\textit{im2txt}} & \multicolumn{10}{c}{\textit{txt2im}}\\
 & \multicolumn{2}{c|}{\textit{Joint}} & \multicolumn{8}{c||}{\textit{Continual}} & \multicolumn{2}{c|}{\textit{Joint}} & \multicolumn{8}{c}{\textit{Continual}}\\
 & \multicolumn{2}{c|}{\textit{CTNP}} & \multicolumn{3}{c|}{\textit{reindexing}} & \multicolumn{5}{c||}{\textit{no reindexing}} & \multicolumn{2}{c|}{\textit{CTNP}} & \multicolumn{3}{c|}{\textit{reindexing}} & \multicolumn{5}{c}{\textit{no reindexing}} \\
 & Yes &  No &  \textit{ft} &  \textit{EWC} &  \textit{MAS} &  \textit{ft} &   \textit{EWC} &  \textit{EWC-im} &  \textit{MAS} &  \textit{MAS-im} & Yes &  No &  \textit{ft} &  \textit{EWC} &  \textit{MAS} &  \textit{ft} &  \textit{EWC} &  \textit{EWC-txt} &  \textit{MAS} &  \textit{MAS-txt} \\  \hline
& \multicolumn{20}{c}{Architecture: \textit{no sharing}} \\
animals & 29.1 & 26.0 & 16.1 & 16.8 & 16.9 & 24.5 & 24.6 & 24.2 & 24.7 & 24.3 
& 27.8 & 25.9 & 15.4 & 15.2 & 15.4 & 20.8 & 20.8 & 20.9 & 19.8 & 20.7 \\
vehicles & 30.9 & 27.7 & 20.8 & 23.3 & 22.7 & 24.0 & 25.1 & 24.8 & 26.0 & 24.8 
& 30.9 & 27.0 & 17.5 & 18.6 & 19.5 & 27.2 & 29.4 & 28.0 & 28.8 & 28.7 \\
clothes & 27.9 & 27.5 & 27.4 & 27.0 & 27.5 & 27.4 & 27.0 & 27.3 & 27.5 & 26.3 
& 29.3 & 27.7 & 28.1 & 27.5 & 28.0 & 28.1 & 27.5 & 27.4 & 28.0 & 28.5 \\ \hline
average & \textbf{\color{green}29.3} & 27.0 & 21.5 & 22.3 & \textbf{22.4} & 24.5 & 24.6 & 24.2 & \textbf{\color{red}24.7} & 24.3
& \textbf{\color{green}29.3} & 26.8 & 20.3 & 20.5 & \textbf{21.0} & 25.4 & 25.9 & 25.4 & 25.6 & \textbf{26.0} \\
A+V+C & \textbf{\color{green}28.5} & 24.4 & 17.0 & \textbf{18.4} & 17.8 & 18.6 & 17.9 & 17.5 & \textbf{\color{red}19.0} & 18.3
& \textbf{\color{green}28.0} & 23.8 & 16.3 & 16.3 & \textbf{16.9} & 20.7 & 21.3 & 20.9 & 20.9 & \textbf{21.4} \\ 
\hline
 & \multicolumn{20}{c}{Architecture: \textit{sharing}} \\
animals & 28.3 & 25.3 & 18.4 & 17.1 & 16.4 & 23.1 & 21.2 & 21.4 & 21.1 & 21.4
& 26.8 & 24.4 & 16.6 & 14.8 & 14.3 & 22.1 & 20.7 & 21.1 & 20.6 & 22.2 \\
vehicles & 30.2 & 28.6 & 22.6 & 24.7 & 23.5 & 23.0 & 24.9 & 25.0 & 23.8 & 26.0
& 31.2 & 27.9 & 16.9 & 17.8 & 16.3 & 27.3 & 29.4 & 29.5 & 28.4 & 28.7 \\
clothes & 26.7 & 27.4 & 27.7 & 26.9 & 27.1 & 27.7 & 26.9 & 27.3 & 27.1 & 26.7
& 27.5 & 26.8 & 27.2 & 27.0 & 26.0 & 27.2 & 27.0 & 27.5 & 26.0 & 28.0 \\ \hline
average & 28.4 & 27.1 & 22.9 & \textbf{22.9} & 22.3 & 24.6 & 24.3 & 24.6 & 24.0 & \textbf{\color{red}24.7}
& 28.5 & 26.4 & \textbf{20.3} & 19.9 & 18.9 & 25.6 & 25.7 & 26.0 & 25.0 &  \textbf{\color{red}26.3}\\
A+V+C & 27.8 & 24.5 & 18.2 & \textbf{18.2} & 17.6 & \textbf{\color{red}19.0} & 17.9 & 18.2 & 17.9 & 18.8
& 27.2 & 23.7 & \textbf{15.9} & 15.5 & 14.9 & 21.8 & 21.5 & 22.2 & 21.0 & \textbf{\color{red}22.6} \\ \hline

\end{tabular}
}
\caption{Results in SeViGe after learning all tasks (Recall@10 in \%). \textit{average} measures performance with \textit{known} task, while \textit{A+V+C} with \textit{unknown} task. Best joint learning result in {\color{green}\textbf{green}}, best continual learning result in {\color{red}\textbf{red}}.}\label{tab:seqvisualgenome}
\end{table*}

\begin{table*}[t]
\centering
\def\arraystretch{1.0}
\setlength\tabcolsep{2pt}
\resizebox{\textwidth}{!}{%
\begin{tabular}{c||cc|ccc|ccccc||cc|ccc|ccccc}
\hline
\multirow{4}{*}{Domain} & \multicolumn{10}{c||}{\textit{im2txt}} & \multicolumn{10}{c}{\textit{txt2im}}\\
 & \multicolumn{2}{c|}{\textit{Joint}} & \multicolumn{8}{c||}{\textit{Continual}} & \multicolumn{2}{c|}{\textit{Joint}} & \multicolumn{8}{c}{\textit{Continual}}\\
 & \multicolumn{2}{c|}{\textit{CTNP}} & \multicolumn{3}{c|}{\textit{reindexing}} & \multicolumn{5}{c||}{\textit{no reindexing}} & \multicolumn{2}{c|}{\textit{CTNP}} & \multicolumn{3}{c|}{\textit{reindexing}} & \multicolumn{5}{c}{\textit{no reindexing}} \\
 & Yes &  No &  \textit{ft} &  \textit{EWC} &  \textit{MAS} &  \textit{ft} &   \textit{EWC} &  \textit{EWC-im} &  \textit{MAS} &  \textit{MAS-im} & Yes &  No &  \textit{ft} &  \textit{EWC} &  \textit{MAS} &  \textit{ft} &  \textit{EWC} &  \textit{EWC-txt} &  \textit{MAS} &  \textit{MAS-txt} \\  \hline
& \multicolumn{20}{c}{Architecture: \textit{no sharing}} \\

task1 & 65.7 & 63.8 & 33.6 & 32.0 & 33.0 & 49.8 & 48.1 & 47.2 & 50.5 & 47.1
&69.7 & 68.2 & 40.1 & 38.0 & 38.2 & 59.8 & 59.2 & 58.3 & 60.0 & 59.7 \\

task2 & 56.5 & 54.9 & 39.8 & 38.5 & 40.0 & 47.0 & 46.6 & 46.4 & 47.0 & 46.9 
&65.2 & 62.6 & 46.8 & 44.7 & 46.9 & 54.6 & 55.5 & 55.1 & 55.5 & 55.9  \\

task3 & 38.2 & 39.9 & 39.7 & 40.1 & 40.2 & 39.7 & 40.1 & 39.9 & 40.5 & 39.7 
&44.6 & 45.7 & 46.7 & 46.7 & 46.0 & 46.7 & 46.7 & 46.7 & 46.0 & 46.2 \\ \hline

average & \textbf{\color{green}53.5} & 52.9 & \textbf{37.7} & 36.9 & \textbf{37.7} & 45.5 & 44.9 & 44.5 & \textbf{\color{red}46.0} & 44.6
& 59.8 & 58.9 & \textbf{44.5} & 43.1 & 43.7 & 53.7 & 53.8 & 53.4 & 53.8 & \textbf{\color{red}54.0} \\

total & \textbf{\color{green}52.4} & 49.8 & \textbf{33.0} & 32.1 & \textbf{33.0} & 37.1 & 36.2 & 35.6 & \textbf{\color{red}37.4} & 36.0
& 58.5 & 56.3 & \textbf{40.4} & 38.7 & 39.7 & 48.3 & 48.0 & 47.3 & 48.2 & \textbf{\color{red}48.4} \\ 
\hline

 & \multicolumn{20}{c}{Architecture: \textit{sharing}} \\

task1 & 65.3 & 63.9 & 32.9 & 31.9 & 34.1 & 48.4 & 47.7 & 47.7 & 47.8 & 45.1
& 70.2 & 67.7 & 38.2 & 37.4 & 39.8 & 58.6 & 56.3 & 58.4 & 57.1 & 57.5 \\

task2 & 55.7 & 55.3 & 40.6 & 39.9 & 40.4 & 46.3 & 46.0 & 45.2 & 44.0 & 44.4
& 64.7 & 63.1 & 46.0 & 45.7 & 46.3 & 54.6 & 54.2 & 55.6 & 54.6 & 54.9 \\

task3 & 37.6 & 40.1 & 39.6 & 39.7 & 39.3 & 39.6 & 39.7 & 39.9 & 40.0 & 39.7
& 44.8 & 46.5 & 46.2 & 45.8 & 45.7 & 46.2 & 45.8 & 45.7 & 46.7 & 46.1 \\ \hline

average & 52.9 & 53.1 & 37.7 & 37.2 & \textbf{37.9} & \textbf{44.8} & 44.5 & 44.3 & 43.9 & 43.1
& \textbf{\color{green}59.9} & 59.1 & 43.5 & 43.0 & \textbf{43.9} & 53.1 & 52.1 & \textbf{53.2} & 52.8 & 52.8\\

total & 51.8 & 50.1 & 33.2 & 32.5 & \textbf{33.5} & \textbf{36.1} & 35.9 & 35.4 & 35.5 & 35.3
& \textbf{\color{green}58.7} & 56.4 & 39.3 & 38.9 & \textbf{39.9} & 47.7 & 46.8 & \textbf{48.1} & 47.1 & 47.5 \\ \hline
\end{tabular}
}
\caption{Results in SeCOCO after learning all tasks (Recall@10 in \%). \textit{average} measures performance with \textit{known} task, while \textit{total} with \textit{unknown} task. Best joint learning result in {\color{green}\textbf{green}}, best continual learning result in {\color{red}\textbf{red}}.}\label{tab:secoco}
\end{table*}

\subsection{Sequential MS-COCO}
\minisection{Sequential MS-COCO (SeCOCO) dataset.} We created a second dataset with image-description pairs of MS-COCO\cite{lin2014microsoft}. Each image in MS-COCO is annotated with five image-level descriptions of the scene and a variable number of object annotations localized to specific regions and labeled with one of 80 disjoint object categories. Object categories are further organized in 12 disjoint super-categories. Organizing the data into tasks is challenging in this case since we want to avoid overlap between tasks, but there are many object annotations in each image. We organized the data into groups of super-categories and removed the images with object annotations in more than one group. After removing overlapping images we use those groups as tasks. We finally selected \textit{animal}, \textit{accessory}, \textit{kitchen}, \textit{food} and \textit{furniture} for task 1, \textit{vehicle}, \textit{outdoor}, \textit{electronic}, \textit{appliance} and \textit{indoor} for task 2 and \textit{person} and \textit{sports} for task 3, with 22475, 13903 and 13919 training images respectively, in addition to 1000/1000 images for validation/test for each task. Note that many other objects and concepts remain unannotated, so there is still semantic overlap across tasks that we cannot control.

\minisection{Cross-modal retrieval.} Table~\ref{tab:secoco} shows the recall@10 for different methods on SeCOCO. In this case joint training also performs better than continual learning methods. The drop due to not training with CTNPs is relatively lower than in SeViSe. This can be explained by a higher semantic overlap between tasks that makes CTNPs less critical. The relative importance of sharing layers is also less important in this case, with very little difference in the results.

Regarding continual learning methods, no reindexing is again the most helpful tool to prevent forgetting. Comparing with the \textit{ft} baselines it gives important boosts of roughly 7-9\%/3-8\% in known/unknown tasks, for both sharing and not sharing layers. As in joint training, sharing layers does not have significant impact in this dataset. Similarly, weight regularization only brings marginal gains. In total, the best result for \textit{txt2im} retrieval improves 9.5\%/8\% over the baseline for known/unknown tasks. For \textit{im2txt} retrieval the improvement is 8.3\%/4.4\%. The results are still far from joint training, so there is space for improvement in future works.

\section{Conclusion}
In this paper we propose, to our knowledge, the first study on how forgetting affects multimodal embedding spaces, focusing on cross-modal retrieval. We propose a continual cross-modal retrieval model that emphasizes the important role of the indexing stage. Cross-modal drifts are also key factors in forgetting in cross-modal tasks. We evaluated several specific tools to alleviate forgetting. 

\section*{Acknowledgements}
We acknowledge the support from Huawei Kirin Solution, the Spanish Government funding for projects PID2019-104174GB-I00 and RTI2018-102285-A-I00, and Kai acknowledges the Chinese Scholarship Council (CSC) No.201706170035. Luis acknowledges the Ramón y Cajal fellowship RYC2019-027020-I.

{\small
\bibliographystyle{ieee_fullname}
\bibliography{egbib}
}

\end{document}